# Enhancing Trustworthy Clinical Diagnosis Decision-Making in Large Language Models via Etiology-Aware Attention Supervision

Peixian Li[a,1], Ruiqi Tu[a], Chengkai Wu[b], Jingjing Ren[c], Yu Tian[a,*], Jingsong Li [a,b,*]

[a] *Engineering Research Center of EMR and Intelligent Expert System, Ministry of Education, Key Laboratory for Biomedical Engineering of Ministry of Education, College of Biomedical Engineering and Instrument Science, Zhejiang University, Hangzhou, 310027, China.*

[b] *School of Intelligent Science and Technology, Hangzhou Institute for Advanced Study, University of Chinese Academy of Sciences, Hangzhou, 310024, China.*

[c] *General Practice Department, the First Affiliated Hospital, College of Medicine, Zhejiang University, Hangzhou 310003, China*

[*] **Corresponding Author:**

Jingsong Li and Yu Tian

College of Biomedical Engineering and Instrument Science, Zhejiang University,

No. 38 Zheda Road, Hangzhou, Zhejiang, 310027, China

Email: ljs@zju.edu.cn (Jingsong Li); tyler@zju.edu.cn (Yu Tian)

## Abstract

Objective: Large Language Models (LLMs) have demonstrated strong capabilities in medical text understanding and generation. However, their trustworthiness in diagnosis-oriented medical tasks remains constrained by the lack of structured guidance on how clinically relevant diagnostic evidence is internally attended to and utilized during model learning.

Method: We propose an Etiology-Aware Attention Supervision framework that introduces structured etiological information as an external supervisory signal for training large language models. Specifically, we construct Clinical Etiology Schema (CES) derived from authoritative clinical guidelines for three acute abdominal conditions: acute appendicitis, acute pancreatitis, and acute cholecystitis. Based on CES annotations, we develop an Etiology-Aware Head Identification strategy to identify attention heads that consistently align with etiological evidence. Building on this analysis, we design a structure-guided parameter-efficient fine-tuning approach that steers attention distributions toward clinically relevant evidence through an additional supervision loss, without modifying the base model architecture.

Result: Experiments conducted on a Consistent Diagnosis Cohort demonstrate that the proposed framework improves average diagnostic accuracy by 15.65% compared with baseline models. Attention-based metrics, including Inference Focus Score and Inference Attention Frequency, show more concentrated attention on etiologically relevant evidence. External evaluation on a Discrepant Diagnosis Cohort further confirms the robustness of diagnostic performance improvements under real-world clinical inconsistencies.

Conclusion: This study presents a practical framework for incorporating clinically grounded diagnostic structures into the training of LLMs, with the explicit goal of enhancing trustworthy clinical diagnostic decision-making. Validated in the context of acute abdominal diagnosis, the results further suggest that etiology-aware attention supervision promotes more structured, clinically consistent, and inspectable evidence utilization, thereby supporting robust and trustworthy diagnostic performance in domain-specific medical applications.

**Keywords—Large Language Model, Trustworthy Clinical Diagnosis, Attention Supervision, Structure-guided Learning, Parameter-Efficient Fine-tuning**

## 1. Introduction

With the rapid advancement of large language models (LLMs), their application in supporting medical decision-making has attracted increasing attention. Pretrained on vast amounts of biomedical literature and clinical records, LLMs have demonstrated strong capabilities in processing unstructured text, answering clinical questions, and generating diagnostic hypotheses (Singhal et al., 2023). Prior studies have explored their potential utility in a range of medical tasks, including triage, clinical report generation, and symptom recognition (Williams et al., 2024).

Despite these advances, LLMs still face notable limitations in real-world clinical applications. In particular, they often struggle to comprehensively collect and integrate critical clinical information and to follow guideline-consistent diagnostic structures when forming predictions (Hager et al., 2024). In contrast, physicians rely on a combination of history-taking, physical examination, laboratory tests, and radiology report to establish accurate diagnoses (Bhangu et al., 2015; Lankisch et al., 2015; Gallaher & Charles, 2022). Accordingly, developing diagnostic models that can reliably and transparently integrate clinical information while adhering to clinically grounded diagnostic structures remains an important challenge for trustworthy clinical decision-making.

Recent research has sought to enhance the clinical performance of LLMs through various strategies. For instance, Kwon et al. (2024) formalized diagnostic processes into chain-of-thought prompts to facilitate clinical performance by encouraging rationale generation; Gao et al. (2025) incorporated structured knowledge graph paths into LLMs' workflows to improve diagnostic accuracy; Hua et al. (2024) fine-tuned LLMs on knowledge-oriented datasets derived from traditional Chinese medicine to strengthen domain-specific knowledge encoding. While these approaches have shown progress in improving interpretability and task performance in specific clinical settings, most rely on prompt engineering or external augmentation and do not directly optimize the internal inference process of LLMs. This limitation becomes particularly salient in complex diagnostic scenarios, where properly guided models have been shown to yield substantial performance gains (Williams, Zack, et al., 2024; Shah-Mohammadi & Finkelstein, 2024). More recently, Fang et al. (2025) demonstrated that optimizing internal training dynamics using model-identified key information can further enhance performance in prognostic prediction tasks, suggesting the promise of internal-level intervention strategies. As a result, such approaches often improve output-level correctness without providing guarantees on how diagnostic evidence is internally utilized, limiting their ability to support trustworthy clinical decision-making.

From a modeling perspective, guiding how internal representations attend to clinically relevant information during inference may offer a more robust and scalable pathway for improving LLM performance on complex medical tasks than approaches operating solely at the input–output level. Given the transformer-based architecture at the core of LLMs—where multi-head self-attention mechanisms capture contextual relationships and feed-forward layers store learned knowledge—targeting the internal attention heads provides a promising opportunity for internal structure-guided (Vaswani et al., 2017; Zheng et al., 2025). Empirical studies have shown that specific attention heads play critical roles in inference and information retrieval. Wu et al. (2024) proposed the Needle-in-a-Haystack test to identify Retriever Heads and demonstrated their utility in reducing hallucination and improving inference consistency. Xiao et al. (2024) explored gating mechanisms to assess the impact of pruning specific heads on performance. Building on these insights, recent work has explored targeted manipulation of attention heads to enhance the reliability and task performance of the model. For example, Huang et al. (2025) employed masked-head control tokens to enhance output faithfulness, while Zhang et al. (2025) improved performance on mathematical tasks through additional calibration of function-specific attention heads. Recent work has explored targeted manipulation of attention heads to enhance the reliability, interpretability, and process-level trustworthiness of model inference.

Motivated by these findings, we propose an Etiology-Aware Attention Supervision framework for structure-guided learning in LLM-based medical diagnosis. We focus on three clinically similar and frequently misdiagnosed acute abdominal emergencies—acute appendicitis, acute pancreatitis, and acute cholecystitis—and construct a Clinical Etiology Schema (CES) based on authoritative clinical guidelines. Patient records are annotated according to CES to embed clinically grounded etiological elements and to provide structural cues for internal attention allocation during training. We further develop an Etiology-Aware Head Identification algorithm to detect attention heads that exhibit strong alignment with etiological element recognition, thereby identifying effective intervention points within the model. During fine-tuning, a structure-guided learning mechanism is applied to steer these heads toward the annotated etiological elements, encouraging the model to follow clinically valid diagnostic structures during inference.

The main contributions of this work are summarized as follows:

1) Etiology-aware attention localization. We introduce an Etiology-Aware Head Identification algorithm that enables precise localization of internal attention heads associated with etiological element recognition, providing a novel internal intervention pathway for improving diagnostic trustworthiness in LLMs, through mechanistically grounded and inspectable attention behaviors.
2) Structure-guided learning and evaluation framework. We propose a structure-guided learning strategy that integrates input-level structural cue embedding with targeted attention supervision, and further establish attention-based evaluation metrics to form a unified pipeline for etiology-aware, structure-guided fine-tuning and assessment of large language models.

This paper first reviews related work on medical LLMs and parameter-efficient fine-tuning, then introduces the Etiology-Aware Attention Supervision framework. Experimental evaluations on two clinical cohorts are subsequently presented, followed by discussion and conclusions.

## 2. Related work

### *2.1. Enhancing Medical Capabilities of LLMs*

Substantial research efforts have focused on enhancing the medical capabilities of LLMs, aiming to improve their diagnostic accuracy, information utilization, and robustness in real-world clinical settings. Existing approaches can be broadly categorized into three directions: prompt-based guidance, internal attention and latent space modulation, and medical knowledge integration.

**Prompt-based Guidance:** Prompt engineering is one of the most widely adopted strategies for improving the medical performance of LLMs. By structuring task instructions and response formats, prompts can shape how models interpret clinical inputs and organize their outputs. Chain-of-thought (CoT) prompting has been shown to encourage stepwise analytical generation and improve transparency in medical tasks (Wei et al., 2022). For example, Savage et al. (2024) demonstrated that CoT-style prompts enabled GPT-4 to produce diagnostically coherent explanations resembling physician narratives. Sonoda et al. (2024) further introduced structured prompt templates that explicitly separate summarization and diagnostic phases, yielding measurable gains in diagnostic accuracy. Wu et al. (2024) proposed a context-filling framework that injects relevant medical concepts retrieved from knowledge graphs into prompts, thereby improving both factual correctness and interpretability. Despite their effectiveness, prompt-based methods operate primarily at the input–output interface and do not directly alter the model's internal representations, limiting their ability to induce stable and transferable diagnostic behaviors.

**Attention and Latent Space Modulation :** To move beyond surface-level guidance, recent studies have explored manipulating internal mechanisms of LLMs to enhance task-specific capabilities. Venkateswaran & Contractor (2025) proposed SpotLight, which dynamically amplifies attention weights for task-relevant tokens, enabling more faithful instruction following. Shen et al. (2025) introduced the Heima framework, which decodes signals from latent representations associated with specific tokens, demonstrating that intermediate hidden states encode task-relevant semantic structure. Hao et al. (2024) extended this idea by feeding selected latent states back into the generation process, allowing continuous refinement of inference trajectories. These works highlight the promise of attention and latent space modulation as a means to improve model behavior at an internal level. However, most existing approaches remain task-agnostic and are not explicitly grounded in domain-specific medical structures, limiting their reliability in clinical applications.

**Medical Knowledge Integration：** Another line of research focuses on enhancing LLM performance through explicit integration of medical knowledge. Parameter updates using curated medical corpora and reasoning-augmented datasets have been shown to improve diagnostic consistency and domain alignment. For instance, MedReason constructs reasoning-chain datasets by combining medical knowledge graphs with public QA resources, enabling models to generate more structured and interpretable diagnostic outputs (Wu et al., 2025). Retrieval-augmented generation methods further incorporate authoritative sources such as clinical guidelines during inference, promoting evidence-based responses (Kresevic et al. 2024). Domain-specific pretraining efforts, including MedFound, have demonstrated that embedding expert-annotated medical knowledge into training data can improve standardization and cross-specialty consistency in diagnosis (Liu et al., 2025). While effective, these methods often treat knowledge as external or static signals and do not explicitly regulate how models internally attend to clinically salient information during decision-making.

Different from prior approaches, our method bridges structured medical knowledge and internal model mechanisms by introducing structure-guided attention supervision during fine-tuning. Rather than solely improving outputs or injecting external knowledge, we identify attention heads consistently aligned with clinically grounded etiological elements and explicitly guide their behavior during training.

*2.2. Parameter-Efficient Fine-Tuning*

As the parameter scale of LLMs continues to grow, full-parameter fine-tuning has become increasingly computationally expensive and often impractical in resource-constrained or high-risk domains such as medicine. To address this challenge, a series of Parameter-Efficient Fine-Tuning (PEFT) methods have been proposed, among which Low-Rank Adaptation (LoRA) and its variants have emerged as particularly effective strategies (Ding et al., 2023; Hu et al., 2021). LoRA, along with variants like QLoRA and AdaLoRA, introduces trainable low-rank matrices into frozen transformer weights (Dettmers et al., 2023; Zhang et al., 2023). By updating less than 1% of parameters, it allows models to adapt to downstream tasks while achieving performance comparable to or better than full-parameter fine-tuning. Building upon this foundation, several studies have proposed enhancements to PEFT techniques to further improve LLMs' performance in the medical domain. For example, Xu et al. (2025) introduced STAF-LLM, which combines AdaLoRA with a task-guided router to selectively integrate knowledge from 12 medical experts. Yan et al. (2025) developed the STP training method, which leverages transformer preference for specific tokens, achieving near full-parameter fine-tuning performance on the GLUE benchmark by updating only 0.009% of the model parameters. In this work, we build upon LoRA by introducing a structure-guided loss that supervises attention distributions during fine-tuning, encouraging the model to consistently attend to clinically relevant information without modifying the full model parameters.

*2.3. Auxiliary Diagnosis Models for Acute Abdomen Emergencies*

Existing studies have highlighted both the potential and limitations of auxiliary diagnostic models in emergency settings. For instance, Males et al. (2024) and Ma et al. (2025) observed that traditional models often fail to produce accurate predictions unless symptom inputs are appropriately structured and interpreted. Hager et al. (2024) developed a benchmark dataset (MIMIC-CDM) consisting of thousands of acute abdomen emergencies. Their evaluations revealed that general-purpose LLMs exhibit suboptimal performance in diagnosing conditions such as acute cholecystitis and acute pancreatitis, primarily due to their limited ability to integrate structured clinical data and adhere to guideline-based reasoning processes. To address these challenges, researchers have proposed innovations tailored to real-world clinical diagnostic workflows. Li et al. (2025) addressed the issue of LLMs overlooking key diagnostic information and lacking confirmatory logic by constructing a stepwise fine-tuning dataset aligned with real clinical workflows (history taking → physical exam → diagnostic tests → final diagnosis), along with a dual-agent diagnostic framework to enforce adherence to clinical diagnostic procedure. To address the limitations of passive clinical input processing, we propose an Etiology-Aware Attention Supervision Framework that explicitly guides LLMs to focus on etiological elements, improving diagnostic discrimination and interpretability in complex emergency scenarios.

## 3. Methods

This study focuses on improving the diagnostic performance of large language models (LLMs) in complex emergency scenarios, with particular emphasis on three clinically confusable acute abdominal emergencies: acute appendicitis, acute pancreatitis, and acute cholecystitis. We propose an Etiology-Aware Attention Supervision Framework that leverages clinically grounded etiological elements to guide internal attention behaviors during model fine-tuning, aiming to enhance trustworthy diagnostic decision-making in high-ambiguity clinical settings, by guiding clinically consistent attention alignment during fine-tuning.

As illustrated in Figure 1, the proposed framework consists of three sequential stages. In the first stage, we construct CES based on authoritative diagnostic guidelines for the target diseases and use it to annotate etiological elements and their relative diagnostic importance within patient records. In the second stage, we develop an Etiology-Aware Head Identification algorithm to identify attention heads whose activation patterns exhibit strong alignment with CES-annotated etiological elements, thereby locating internal attention mechanisms associated with etiological information processing. In the third stage, we integrate a structure-guided attention supervision loss into a fine-tuning process using LoRA. This loss function explicitly encourages the identified etiology-aware attention heads to attend more strongly to CES-annotated elements during training, guiding the model's internal attention allocation toward clinically grounded diagnostic structures.

The following subsections describe each stage of the proposed framework in detail.

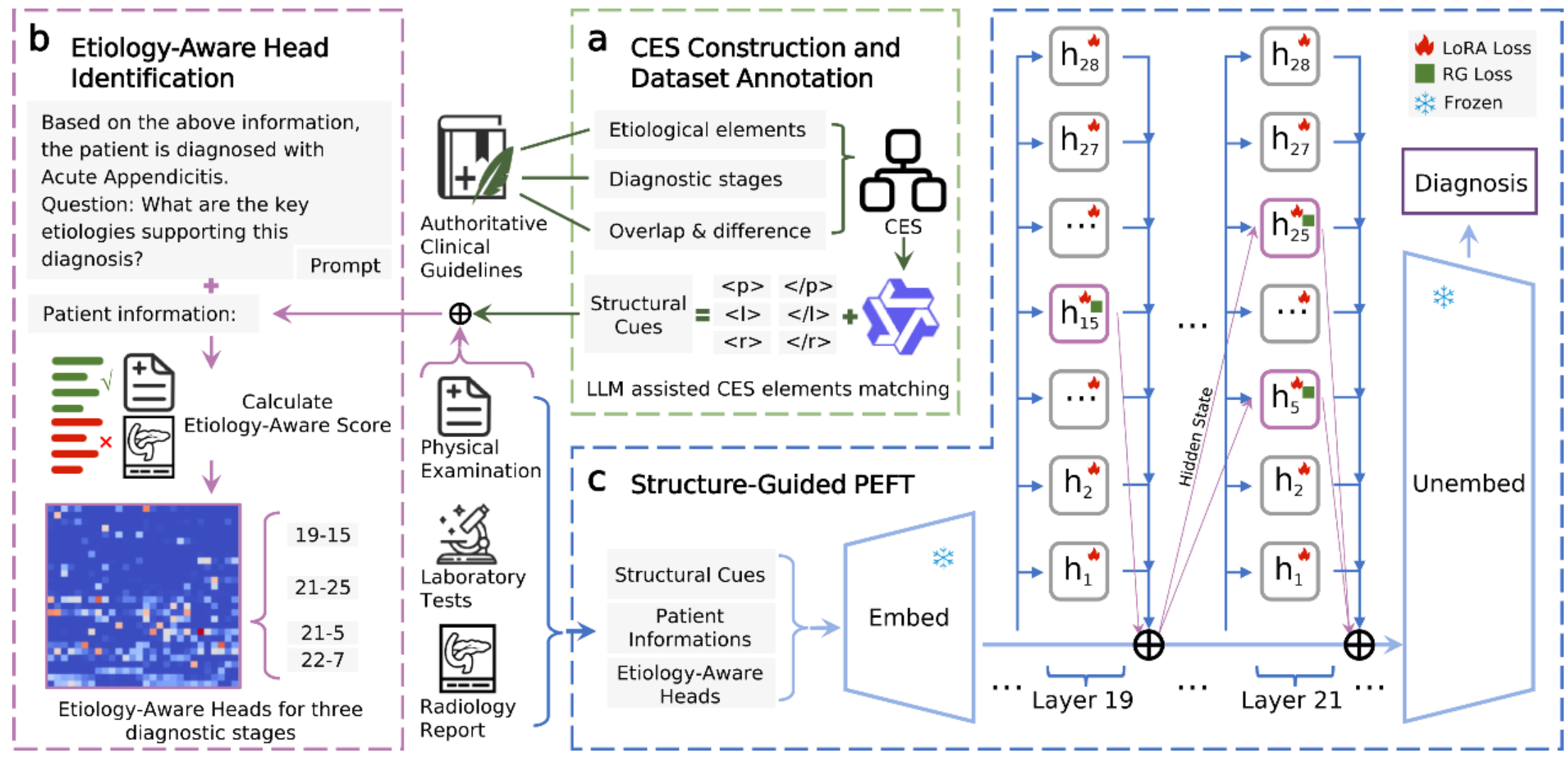

**Figure 1.** Overview of the Etiology-Aware Attention Supervision Framework. (a) Clinical Etiology Schema Construction and Dataset Annotation: Authoritative diagnostic guidelines are used to construct the Clinical Etiology Schema (CES), which is then applied to annotate etiological elements and their diagnostic relevance within patient records. (b) Etiology-Aware Head Identification: Attention heads whose activation patterns exhibit strong alignment with CES-annotated etiological elements are identified, serving as internal attention targets for subsequent supervision. (c) Structure-Guided Parameter-Efficient Fine-Tuning: A structure-guided attention supervision loss is integrated into a LoRA-based PEFT process, encouraging the identified etiology-aware attention heads to focus on CES-annotated elements during training and guiding internal attention allocation toward clinically grounded diagnostic structures.

*3.1. Clinical Etiology Schema Construction and Dataset Annotation*

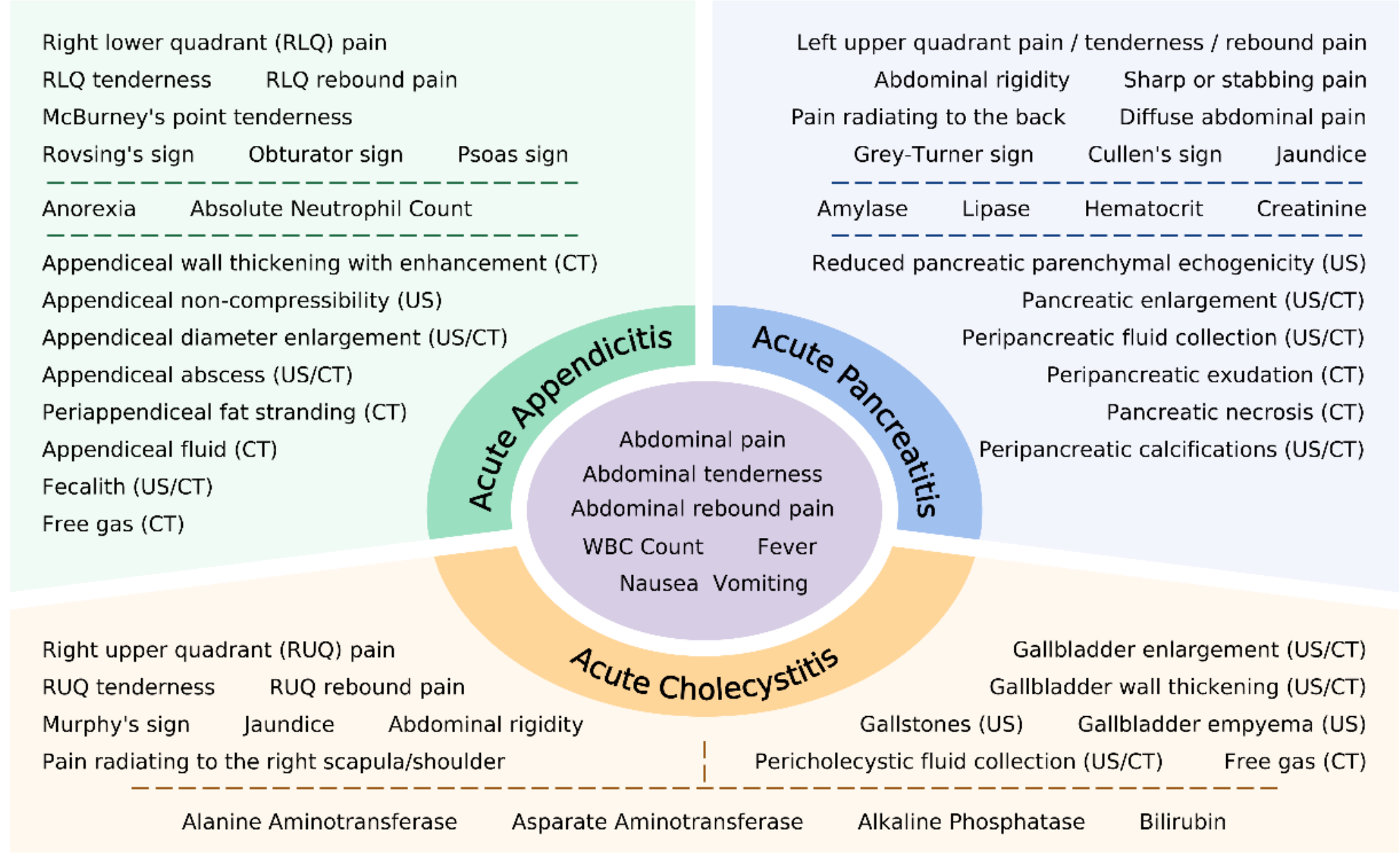

**Figure 2.** Clinical Etiology Schema (CES) derived from authoritative diagnostic guidelines. For the three target diseases, CES organizes etiological elements across physical examination, laboratory tests, and radiology reports, providing a clinically grounded diagnostic structures.

To construct the clinically grounded diagnostic structures, we systematically analyzed authoritative clinical guidelines published in recent years (Kumar et al., 2024; Di Saverio et al., 2020; Moris et al., 2021; Takada et al., 2013, 2007; Chinese Pancreatic Surgery Association, Chinese Society of Surgery, Chinese Medical Association, 2021; Lankisch et al., 2015). We focused on diagnostic procedures for three acute abdominal emergencies—acute appendicitis, acute pancreatitis, and acute cholecystitis—and derived the CES from these guidelines. These guidelines represent international or national expert consensus statements with high clinical recognition and practical relevance, covering standardized diagnostic workflows for the target diseases.

The CES abstracts the real-world diagnostic process into three major evidence categories: physical examination, laboratory tests, and radiology reports. Each category consists of representative etiological elements that are commonly referenced in guideline-based diagnosis. In its design, the CES explicitly emphasizes both shared and disease-specific clinical manifestations across the three conditions, enabling the structured characterization of etiological relevance through complementary diagnostic evidence. The overall structure of the CES is illustrated in Figure 2.

To support subsequent stages of Etiology-Aware Head Identification and structure-guided fine-tuning, patient records were annotated according to the CES. Specifically, we first employed a locally deployed large language model (Qwen2.5-32B-Instruct) with carefully designed prompts (Supplementary Data 1.1) to automatically identify CES-related etiological elements within unstructured clinical narratives. The generated annotations were then manually reviewed and refined to correct conceptual ambiguities and formatting inconsistencies, ensuring annotation accuracy and consistency. To enable precise localization of CES-related information during model training, we introduced structure tokens as explicit markers for different etiological categories. Dedicated start–end token pairs were used for each evidence type: <p> </p> for physical examination, <l> </l> for laboratory tests, and <r> </r> for radiology report. These tokens serve as structural cues that allow the model to distinguish between different types of diagnostic evidence without introducing explicit task-specific labels at inference time.

$$M = \{s_1, s_2, \dots, s_p, \dots, s_l, \dots, s_r, \dots, s_m\} \tag{1}$$

$$s_p = [t_1, t_2, \dots, t_i, \dots, t_j, \dots, t_n] \xrightarrow{note} s_p = [t_1, t_2, \dots, < p >, t_i, \dots, t_j, </p>, \dots, t_n] \tag{2}$$

Formally, given a patient record $M$ consisting of $m$ sentences, if a sentence $s_p$ contains $n$ tokens corresponding to physical examination findings, the annotation process can be represented in vectorized form as shown in Equations (1) and (2), where $t$ denotes an individual token after tokenization. The final annotated record is obtained by inserting the corresponding start and end structure tokens into sentences $s_p$, $s_l$, and $s_r$ that contain CES-related etiological elements.

*3.2. Etiology-Aware Head Identification*

To characterize how different attention heads respond to clinically grounded diagnostic structures, we propose an Etiology-Aware Head Identification procedure that quantifies the alignment between attention distributions and CES-annotated etiological elements. Rather than attributing explicit functional or causal roles to individual heads, this procedure aims to identify attention heads whose activation patterns exhibit consistent statistical association with etiological elements during diagnosis-related generation. The core idea is to construct a task-specific prompting setting in which the model is required to generate diagnostic procedure conditioned on patient records, and then analyze the attention distributions of each head across transformer layers during token-by-token generation. In this context, etiological elements denote CES-annotated span of physical examination, laboratory tests, and radiology reports from the input. For each attention head, we compute an Etiology-Aware Score (EAS) that measures how frequently the head assigns its maximum attention weight to CES-annotated etiological tokens during generation. The detailed procedure is summarized in Algorithm 1.

Specifically, for each diagnostic stage defined in the CES, we process an Etiology-Aware Head Identification dataset $D^I$ using a standardized task-specific prompt (Supplementary Data 1.2). During generation, we extract the attention score matrices corresponding to each head at each transformer layer. For every generation step, we identify the input token receiving the highest attention weight from the given head and record whether this token lies within the CES-annotated span for the corresponding diagnostic stages. The EAS value for a head is obtained by aggregating these counts across samples and normalizing by the number of annotated tokens for the given stage. This results in a matrix $EAS \in \mathbb{R}^{G\times L\times H}$, where $G$ denotes the number of diagnostic stages, $L$ the number of transformer layers, and $H$ the number of attention heads per layer. Attention heads with higher EAS values are interpreted as exhibiting stronger alignment with specific diagnostic stages and are selected as etiology-aware heads for subsequent structure-guided fine-tuning.

---

Algorithm 1 Computation of Etiology-Aware Scores

---

**Input：**

Etiology-Aware Head Identification dataset $D^I$

Task-specific prompt (see Supplementary Data 1.2)

**Output：**

Etiology-Aware score matrix $EAS \in \mathbb{R}^{G\times L\times H}$, where $G$ is number of diagnostic stages (Physical exam, Lab tests, Radiology), $L$ is number of model's transformer layers and $H$ is number of attention heads per layer

**Procedure:**

Initialize $EAS \leftarrow 0 \in \mathbb{R}^{G\times L\times H}$

**for** each stage index $i = 1\ to\ G$ ：

    **for** each layer index $j = 1\ to\ L$ ：

        **for** each head index $k = 1\ to\ H$ ：

            **for** each instance $x \in D^I$ ：

                Feed $x$ and prompt into the model

                Extract the attention score matrix $A_{j,k}$ for the head $(j,k)$

                Identify the token index with the maximum attention score

                **if** this index lies within the CES-labeled etiology token range:

                    Increment $EAS[i][j][k] \leftarrow EAS[i][j][k] + 1$

            Normalize $EAS[i][j][k]$ by dividing by the number of etiology tokens in stage $i$

**return** $EAS$

---

It is important to emphasize that the EAS is not intended to establish causal responsibility of individual attention heads for diagnostic correctness. Instead, EAS functions as a principled selection criterion to identify attention heads that are consistently sensitive to CES-annotated information. This design provides well-defined and interpretable intervention targets for attention supervision in the subsequent training stage.

*3.3. Structure-Guided Parameter-Efficient Fine-Tuning*

To encourage the model to allocate attention to clinically grounded etiological elements during diagnosis generation, we incorporate a structure-guided loss into the parameter-efficient fine-tuning. This auxiliary objective is designed to guide attention behavior toward CES-annotated structural cues while preserving the original diagnostic generation capability of the model.

We adopt Low-Rank Adaptation (LoRA) as the PEFT strategy. Under this setting, the backbone parameters of the pre-trained language model are kept frozen, and only the introduced low-rank matrices are updated during training. As the primary optimization objective, we employ label smoothing cross-entropy loss, which alleviates overconfidence in hard labels by smoothing the target distribution. Specifically, the ground-truth label distribution $y_i$ is transformed into a smoothed distribution $\tilde{y}_i$ using a smoothing factor $\varepsilon$ and the total number of classes $K$, as defined in Equations (3) and (4).

$$\tilde{y}_i = \begin{cases} 1-\varepsilon, & i = true\ lable \\ \dfrac{\varepsilon}{K-1}, & others \end{cases} \tag{3}$$

$$\mathcal{L}_{smooth} = -\sum_i \tilde{y}_i \log p_i \tag{4}$$

On top of the base loss, we introduce the structure-guided loss to quantify the extent to which selected attention heads focus on CES-annotated etiological elements during prediction. For a given attention head $H_{l,h}$ with attention matrix $A_{l,h} \in \mathbb{R}^{n\times n}$, we define the attention allocation score $Att(H_{l,h}, s_t)$ for a target etiological segment $s_t$ as follows:

$$Att(H_{l,h}, s_t) = \frac{\sum_{y=1}^{n} \sum_{x=i}^{j} A_{l,h}(x,y)}{\sum_{y=1}^{n} \sum_{x=i}^{n} A_{l,h}(x,y)} \quad (5)$$

where $n$ denotes the input sequence length, and $i$ and $j$ indicate the start and end token positions of the corresponding CES-annotated etiological elements. This score measures the proportion of attention mass assigned to the target token span relative to the total attention of the head.

During structure-guided fine-tuning, attention allocation scores are computed for all previously identified etiology-aware heads with respect to their corresponding stages. These scores are then incorporated into the overall training objective via a weighted combination with the label smoothing loss, as shown in Equation (6). The weighting coefficient $\lambda$ controls the contribution of the structure-guided loss, while $e$ and $E$ denote the current and total training steps, respectively. To balance guidance and generalization, we apply a cosine decay schedule to gradually reduce the influence of the structure-guided loss over training, assigning stronger guidance in the early stages and attenuating it in later phases.

$$\mathcal{L}_{RG} = \mathcal{L}_{smooth} + \frac{\lambda}{2}(1 + cos\frac{e\pi}{E})[1 - \frac{1}{|H|}\sum_{H} Att(H,s)] \quad (6)$$

Through this training strategy, the model is encouraged to align its internal attention behavior with etiological elements and clinically structural cues, while preserving its general generation capabilities. The proposed framework thus integrates structure-guided learning into the fine-tuning process in a soft and interpretable manner, without imposing hard constraints on model predictions.

## 4. Experiments

### *4.1. Dataset*

This study constructed two distinct datasets for evaluating model performance: the Consistent Diagnosis Cohort and the Discrepant Diagnosis Cohort. The former is designed to assess model performance under clinically complete and internally consistent conditions, while the latter targets more challenging real-world scenarios characterized by diagnostic ambiguity, incomplete information, and initial misclassification.

**Consistent Diagnosis Cohort**

The Consistent Diagnosis Cohort was derived from the MIMIC-IV database (v2.2) (Johnson et al., 2023; Johnson et al., 2021; Goldberger et al., 2000), a large-scale, de-identified electronic health record (EHR) repository developed and maintained by the Massachusetts Institute of Technology (MIT). The database contains inpatient and emergency care records for approximately 300,000 patients treated at Beth Israel Deaconess Medical Center between 2008 and 2019. All authors completed the required ethics training and obtained authorized access to the database. Dataset construction partially followed the preprocessing pipeline proposed by Hager et al. (2024) for MIMIC-CDM to improve data quality and enhance comparability with prior studies. This pipeline includes structured data conversion, clinical text cleaning, and diagnostic cohort selection, with full implementation details publicly available on GitHub and PhysioNet.

The Consistent Diagnosis Cohort focuses on three clinically confusable acute abdominal emergencies: acute appendicitis, acute cholecystitis, and acute pancreatitis. The initial candidate pool consisted of patients whose primary ICD diagnosis at admission corresponded to one of the three target diseases. To ensure clinical completeness, we retained only records containing documentation from all three key diagnostic stages: physical examination, laboratory testing, and radiology reports. Records missing any of these components were excluded. To better reflect first-encounter emergency presentations, we further removed cases in which the discharge summary explicitly mentioned one of the target diseases in the history of present illness. Such records typically correspond to transferred patients or recurrent admissions and do not represent an initial diagnostic setting. In addition, records with discharge diagnoses dominated by major conditions unrelated to the target diseases were excluded. After filtering, the final cohort comprised 957 cases of acute appendicitis, 648 cases of acute cholecystitis, and 538 cases of acute pancreatitis. Detailed demographic and clinical statistics are provided in Supplementary Data 2.

**Discrepant Diagnosis Cohort**

To evaluate model behavior under diagnostically challenging conditions, we independently constructed a Discrepant Diagnosis Cohort focusing on cases with inconsistencies between initial and final diagnoses. This cohort was extracted from outpatient EHR data collected at the First Affiliated Hospital, Zhejiang University School of Medicine, covering the period from 2010 to 2021. The study protocol was approved by the hospital's Clinical Research Ethics Committee (Approval No. 2022-809).

This cohort includes patients presenting with abdominal symptoms whose final confirmed diagnosis was one of the three target diseases, while the initial outpatient diagnosis differed. For example, a patient ultimately diagnosed with acute cholecystitis may have been initially assessed as having acute pancreatitis or gallstone disease. Such cases reflect realistic diagnostic uncertainty and symptom overlap commonly encountered in early clinical decision-making. Using these criteria, the Discrepant Diagnosis Cohort consisted of 103 cases of acute appendicitis, 54 cases of acute cholecystitis, and 106 cases of acute pancreatitis. Demographic and clinical characteristics for this cohort are also summarized in Supplementary Data 2.

*4.2. Experimental Setup*

To support local deployment and enable customized fine-tuning, we selected two open-source large language models: Qwen2.5-7B-Instruct and DeepSeek-R1-Distill-Qwen-7B. These models were used consistently across the Etiology-Aware Head Identification stage, the Structure-Guided Parameter-Efficient Fine-Tuning stage, and the final performance evaluation.

Qwen2.5-7B-Instruct is a general-purpose LLM trained through a multi-stage pipeline (Qwen et al., 2025). It was pretrained on over 18 trillion tokens and subsequently fine-tuned on more than one million high-complexity task samples using a combination of supervised learning and reinforcement learning. The model demonstrates strong performance across a wide range of benchmarks, including language understanding, mathematics, programming, and human preference alignment. DeepSeek-R1-Distill-Qwen-7B is a reasoning-oriented lightweight model distilled from DeepSeek-R1 using the Qwen architecture (DeepSeek-AI et al., 2025). Through knowledge distillation, it preserves much of the teacher model's reasoning capability while substantially reducing computational and deployment costs, making it suitable for local experimentation and clinical-oriented evaluation.

For the Etiology-Aware Head Identification task, we randomly sampled 120 patient records for each of the three target diseases to construct disease-specific subsets, along with an additional 40 mixed cases drawn uniformly from all three disease categories. All identification experiments were conducted on a single NVIDIA RTX 4090 GPU. For the structure-guided fine-tuning, we employed a five-fold cross-validation strategy to ensure robustness and reduce variance in performance estimation. Hyperparameters were tuned to achieve stable and efficient training, and all fine-tuning experiments were performed on a single NVIDIA A100 GPU. The complete experimental codebase and configuration details are publicly available at: https://github.com/Twilight-sp/EAS-Diagnosis.

*4.3. Baseline and Evaluation Metrics*

To validate the effectiveness of the proposed framework, we evaluated multiple baseline and comparison models, which are categorized as follows:

**Qwen/DeepSeek-distill:** the original Qwen2.5-7B-Instruct and DeepSeek-R1-Distill-Qwen-7B models without fine-tuning.

**Qwen(LoRA)/DeepSeek-distill(LoRA):** models fine-tuned using standard LoRA-based fine-tuning.

**Qwen(our)/DeepSeek-distill(our):** models fine-tuned using the proposed Etiology-Aware Attention Supervision framework.

**Llama-70B**: the Llama-3.3-70B-Instruct model, reported as the best-performing model in the diagnostic evaluation conducted by Hager et al. (2024), serving as a strong external reference.

To evaluate diagnostic performance, we adopted standard multi-class classification metrics. Overall accuracy measures the proportion of cases in which the target disease was correctly identified, while per-class accuracy (recall) reflects the proportion of correctly diagnosed cases within each disease category.

To assess inference behavior during diagnosis, we introduced the Inference Focus Score, which quantifies the extent to which the model's attention is concentrated on clinically grounded etiological elements within the input sequence. In addition, we employed ROUGE-L F1, a widely used metric for free-text evaluation, to assess the semantic coherence and information preservation of the model response.

To further characterize how models allocate attention to semantically meaningful input segments, we introduced the Inference

Attention Frequency metric. This segment-level metric counts how often each clinically grounded etiological element is attended to by the model across the entire evaluation dataset and normalizes this count by the total number of occurrences of that segment. The resulting value reflects the consistency with which the model focuses on relevant clinical information during the diagnostic process.

*4.4. Results in Consistent Diagnosis Cohort*

We first conducted Etiology-Aware Head Identification using the Qwen and DeepSeek-distill models on the Consistent Diagnosis Cohort. For each attention head, the EAS was computed across three CES stages. To ensure comparability across diseases, scores were normalized using balanced mixed subsets containing an equal number of cases from each disease category. The top-10 scoring attention heads for each stage are visualized in Figure 3, where each vertical label *l–h* denotes the *h*-th attention head in the *l*-th transformer layer.

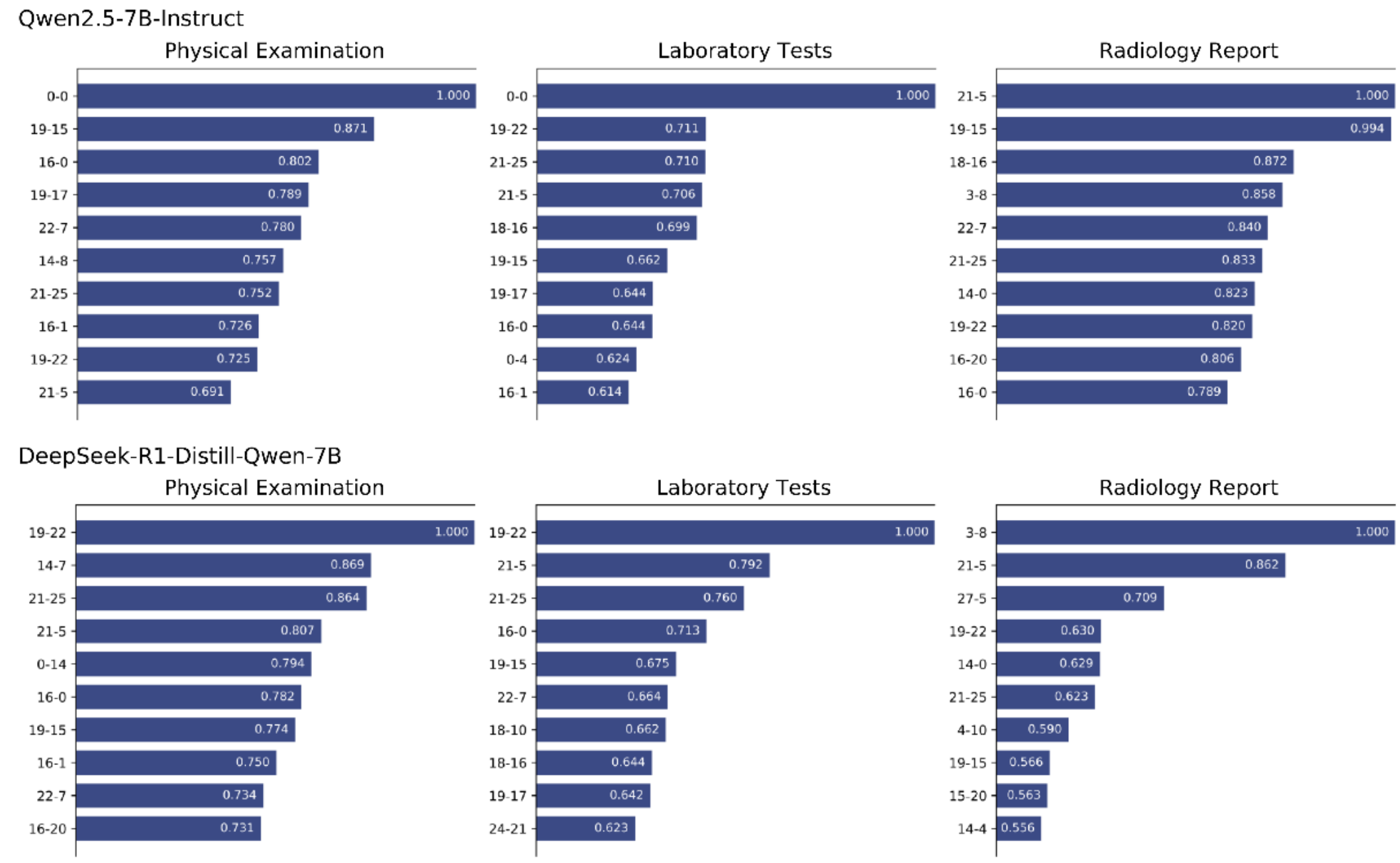

**Figure 3.** Top-10 attention heads ranked by Etiology-Aware Score across three CES stages for both models. Evaluations were conducted on balanced mixed datasets. Scores were normalized. Vertical labels indicate *layer–head* identifiers.

The results indicate that both Qwen and DeepSeek-distill contain multiple attention heads that are highly sensitive to CES-annotated etiological elements across all stages. Moreover, certain heads exhibit cross-stage consistency. For example, in DeepSeek-distill, head *19–22* achieved the highest score in the physical examination stage and also ranked among the top heads in the laboratory and radiology stages, suggesting a stable role in attending to diagnostically relevant information throughout the inference process.

To further examine whether identified attention patterns are consistent across diseases, we computed pairwise Jaccard similarity matrices between the sets of top-10 heads identified separately for each disease. As shown in Figure 4, Qwen demonstrated high inter-disease similarity across stages (mostly >0.8), indicating stable and largely disease-agnostic sensitivity to CES-defined elements. In contrast, DeepSeek-distill exhibited more disease-specific attention patterns, particularly in the laboratory test and radiology stages, where the overlap of top heads across diseases was substantially lower.

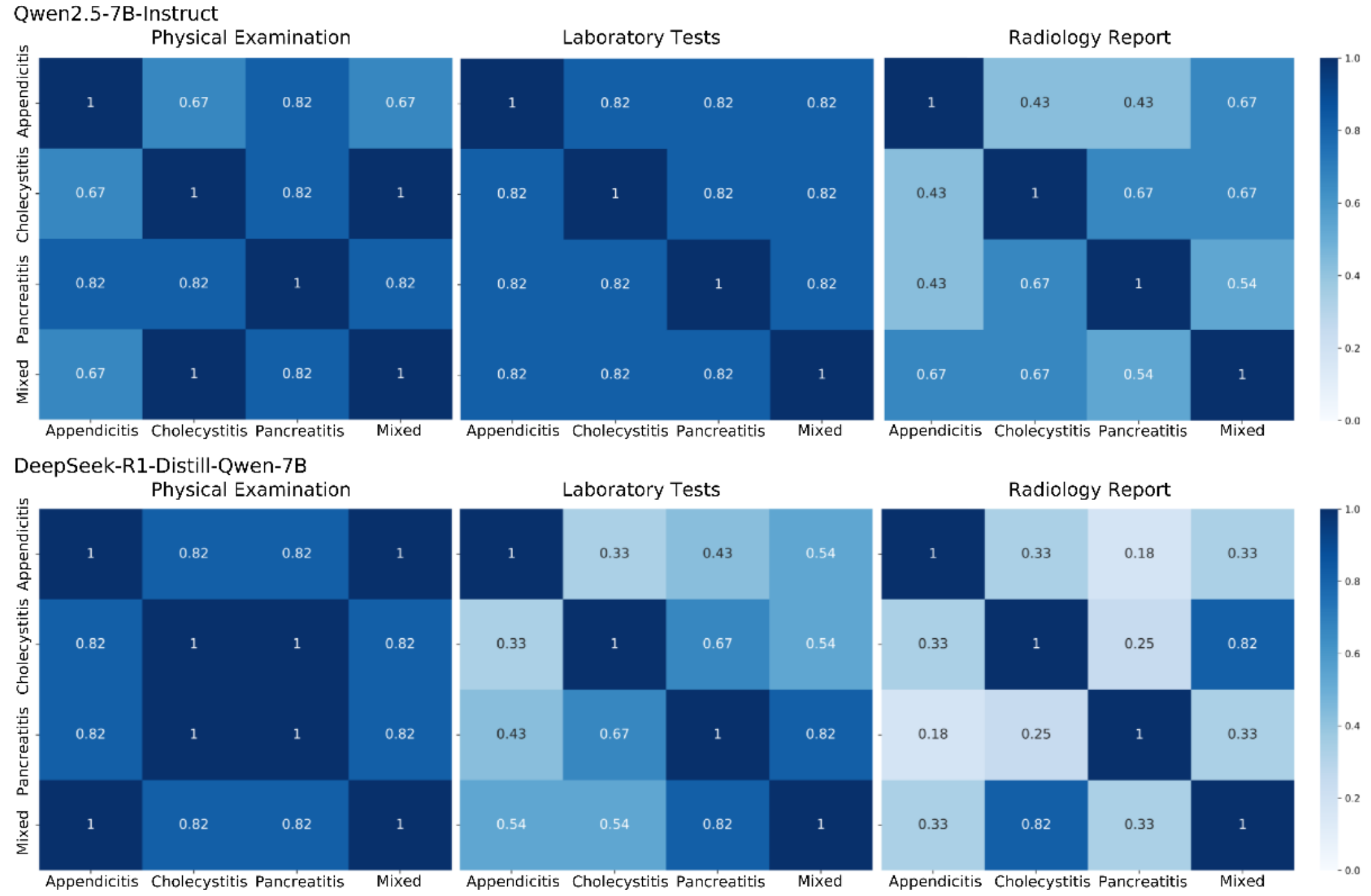

**Figure 4**. Jaccard similarity matrices of top-10 attention heads across disease-specific subsets. Evaluations were conducted on appendicitis-only, cholecystitis-only, pancreatitis-only, and mixed datasets. Scores were normalized prior to similarity computation.

Based on these analyses, we selected a subset of prominent attention heads from each model as etiology-aware heads for subsequent structure-guided fine-tuning. For Qwen, heads *19–15* were selected for the physical examination stage, heads *21–25* for the laboratory test stage, and heads *21–5* and *22–7* for the radiology report stage. For DeepSeek-distill, heads *19–22* were selected for the physical examination stage, heads *21–5* for the laboratory test stage, and heads *3–8* and *27–5* for the radiology report stage.

Both models were then fine-tuned using standard LoRA and the proposed structure-guided strategy. Diagnostic performance was evaluated against their respective base models. Figure 5 presents per-disease accuracy as well as overall accuracy across the three target diseases.

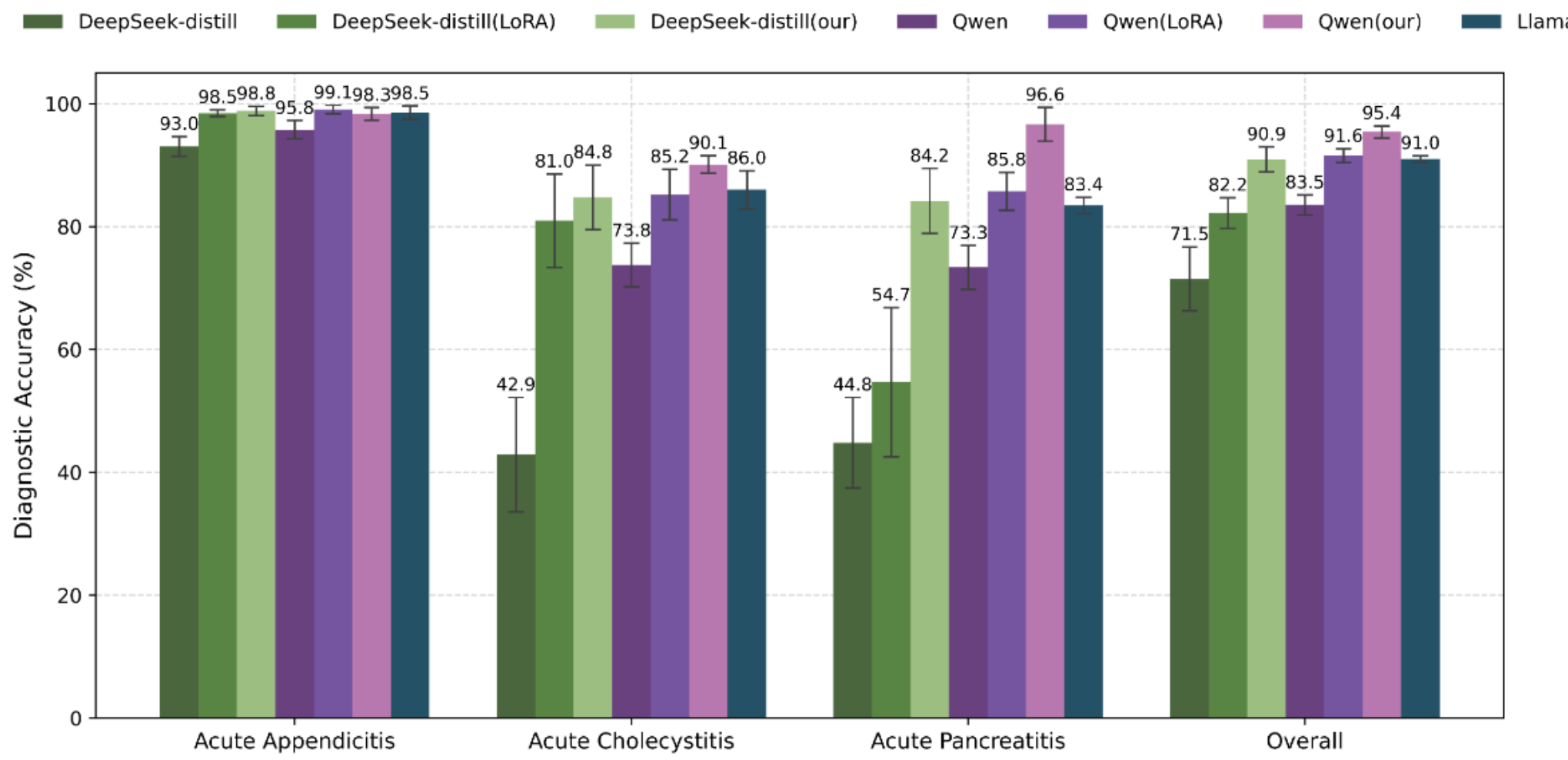

**Figure 5.** Diagnostic accuracy (%) comparison in the Consistent Diagnosis Cohort. Results are shown for base models, LoRA fine-tuned models (LoRA), and structure-guided fine-tuning models (Ours), evaluated both overall and per disease. Error bars indicate 95% confidence intervals derived from five-fold cross-validation using the Z-distribution. Model sizes are as follows: Llama (70B parameters), DeepSeek-distill (7B), and Qwen (7B).

Among base models, Llama achieved the highest accuracy for acute appendicitis (97.9%), but its performance declined for the more clinically confusable conditions of acute cholecystitis (86.0%) and acute pancreatitis (83.4%). Qwen and DeepSeek-distill exhibited further reductions in these tasks, with DeepSeek-distill reaching a minimum accuracy of 42.9% for acute cholecystitis. Applying standard LoRA fine-tuning led to consistent performance improvements across models. Qwen(LoRA) achieved the highest accuracy for acute appendicitis (99.1%), while DeepSeek-distill(LoRA) showed notable gains for acute cholecystitis (81.0%).

When applying the proposed structure-guided strategy, all models demonstrated further improvements across disease-specific and overall diagnostic performance. Notably, Qwen(our) achieved accuracies of 90.1% for acute cholecystitis and 96.6% for acute pancreatitis, corresponding to absolute improvements of 15.7% and 23.3% over Qwen(LoRA), respectively. DeepSeek-distill(our) reached an accuracy of 73.3% for acute pancreatitis, representing a substantial improvement of 29.5% compared to DeepSeek-distill(LoRA).

To assess statistical significance, we conducted one-sided Wilcoxon signed-rank tests. Results revealed statistically significant improvements ($P = 0.0312 < 0.05$) in overall accuracy and per-disease diagnostic accuracy when comparing Qwen(our) with Qwen, and DeepSeek-distill(our) with DeepSeek. Furthermore, when compared with their respective LoRA fine-tuned versions, both Qwen(our) and DeepSeek(our) demonstrated significant gains in overall accuracy and in the tasks of acute cholecystitis and acute pancreatitis ($P = 0.0312$). Detailed results are shown in Supplementary Data 3.

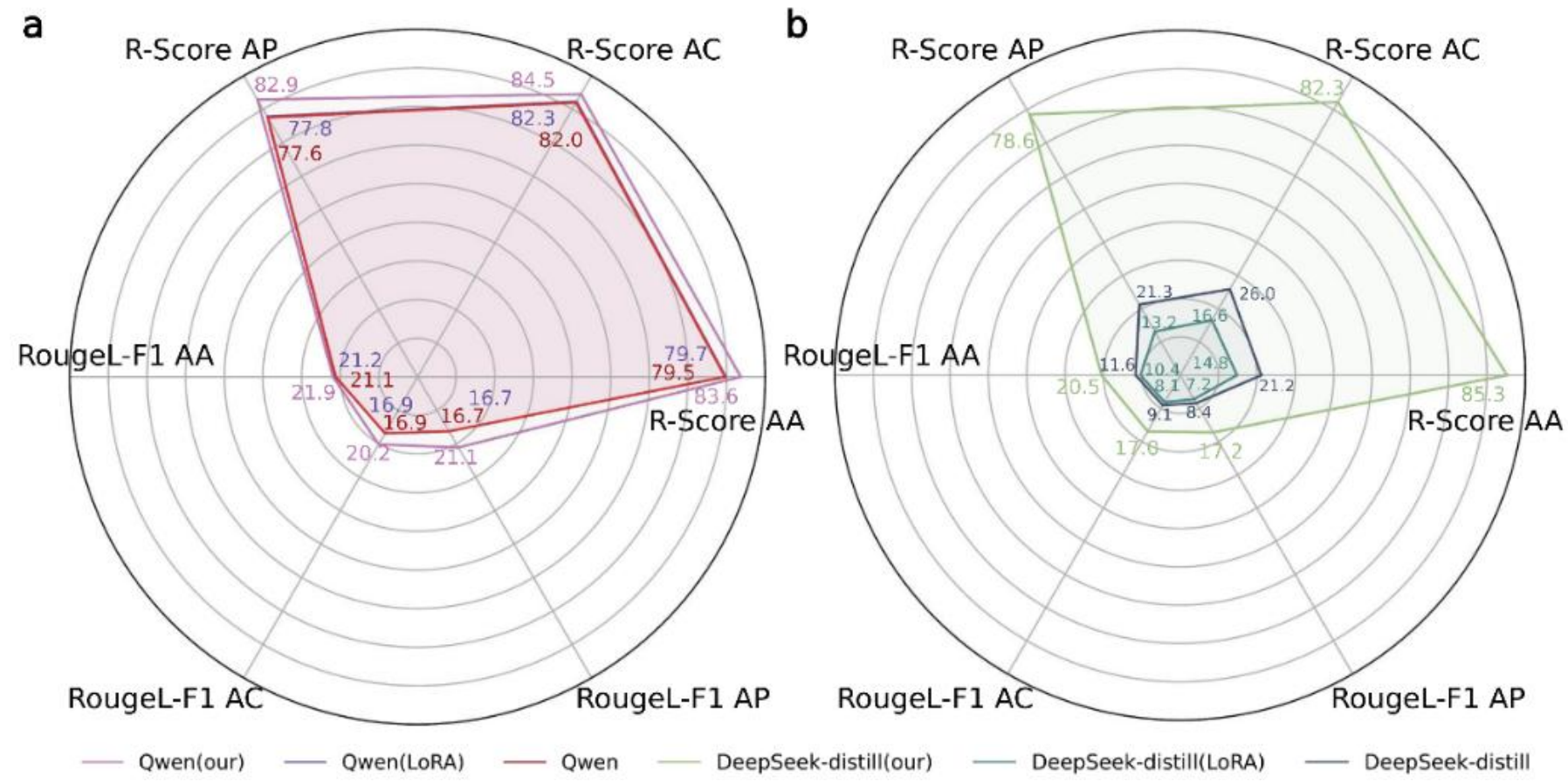

**Figure 6.** Evaluation of inference-related metrics. (a) Comparison among Qwen variants. (b) Comparison among DeepSeek-distill variants. AA: Acute Appendicitis; AC: Acute Cholecystitis; AP: Acute Pancreatitis; R-Score: Inference Focus Score; RougeL-F1: ROUGE-L F1 score.

To further analyze inference behavior during diagnostic generation, we evaluated models using two inference-related metrics. The Inference Focus Score quantifies the extent to which model attention concentrates on CES-annotated etiological elements during decoding. ROUGE-L F1 is applied to the decoded sequence of token representations to measure their alignment with the semantic structure of the target diagnosis. As shown in Figure 6a, Qwen(our) consistently achieved the highest Inference Focus Scores across all three diseases, indicating more concentrated attention on clinically relevant information. In terms of ROUGE-L F1, Qwen(our) outperformed other models in acute cholecystitis and acute pancreatitis, while exhibiting comparable performance to the base model for acute appendicitis. Results for the DeepSeek-distill family (Figure 6b) show that DeepSeek-distill(our) achieved consistent improvements across both metrics. In contrast, DeepSeek-distill(LoRA) underperformed the base model across multiple tasks, suggesting potential misalignment effects when applying standard fine-tuning to reasoning-specialized models.

To further validate the effectiveness of individual components in the proposed framework, we conducted ablation studies focusing on three key modules: CES-based data annotation, Etiology-Aware Head Identification, and the structure-guided loss. Specifically, we evaluated three ablation settings: (1) w/o CES, where annotated structural cues were removed; (2) w/o E-A Head, where the structure-guided loss was applied uniformly to attention heads without head selection; and (3) w/o R-G Loss, where annotated data were retained but the additional loss term was removed.

**Table 1.** Ablation study results in the Consistent Diagnosis Cohort. AA: Acute Appendicitis, AC: Acute Cholecystitis, AP: Acute

Pancreatitis.

| Model | Accuracy | Macro-F1 | AA Recall | AC Recall | AP Recall |
| --- | --- | --- | --- | --- | --- |
| Qwen(our) | **0.9542** | **0.9468** | 0.9833 | 0.9011 | **0.9665** |
| Qwen w/o CES | 0.9156 | 0.8992 | 0.9906 | 0.8521 | 0.8575 |
| Qwen w/o E-A Head | 0.7389 | 0.6827 | 0.9427 | 0.3077 | 0.8972 |
| Qwen w/o R-G Loss | 0.9415 | 0.9290 | **0.9948** | **0.9400** | 0.8431 |
| DeepSeek-distill(our) | **0.9094** | **0.8938** | 0.9885 | 0.8477 | **0.8420** |
| DeepSeek-distill w/o CES | 0.8221 | 0.7807 | 0.9847 | 0.8096 | 0.5466 |
| DeepSeek-distill w/o E-A Head | 0.7415 | 0.6684 | 0.9607 | 0.7414 | 0.3034 |
| DeepSeek-distill w/o R-G Loss | 0.9021 | 0.8783 | **0.9896** | **0.9462** | 0.6915 |

In the w/o CES setting, only label smoothing cross-entropy loss was applied during training, resulting in performance comparable to standard LoRA fine-tuning. In the w/o E-A Head setting, attention heads were manually partitioned by layer depth due to the absence of EAS-based selection, leading to reduced performance gains. In the w/o R-G Loss setting, annotated data were used but no attention supervision was applied, yielding intermediate results between LoRA and the full model. Across all ablation settings, all other model architectures and training configurations were held constant.

*4.5. Results in Discrepant Diagnosis Cohort*

We evaluated the performance of both Qwen and DeepSeek-distill models on the Discrepant Diagnosis Cohort, along with their models fine-tuned on the Consistent Diagnosis Cohort using either LoRA or our method. The results are shown in Figure 7. As expected, diagnostic accuracy decreased across all models compared to the Consistent Diagnosis Cohort, reflecting the increased difficulty of clinical decision-making under conditions characterized by ambiguous symptom presentations and incomplete or misleading clinical context.

The base Qwen model achieved an accuracy of 70.4%, while Qwen(LoRA) exhibited a performance drop to 67.0%. In contrast, Qwen(our) achieved the highest accuracy of 76.6%, representing a clear improvement over both the base and LoRA fine-tuned models. The DeepSeek-distill family demonstrated lower overall performance on this cohort. The base DeepSeek-distill model achieved an accuracy of 32.8%, which further declined to 28.1% after LoRA fine-tuning. In contrast, our method substantially improved performance to 42.2%, yielding an absolute gain of nearly 10 percentage points over the base model. This pronounced improvement suggests that structure-guided fine-tuning is particularly effective in enhancing robustness for smaller models when confronted with ambiguous or conflicting diagnostic cues.

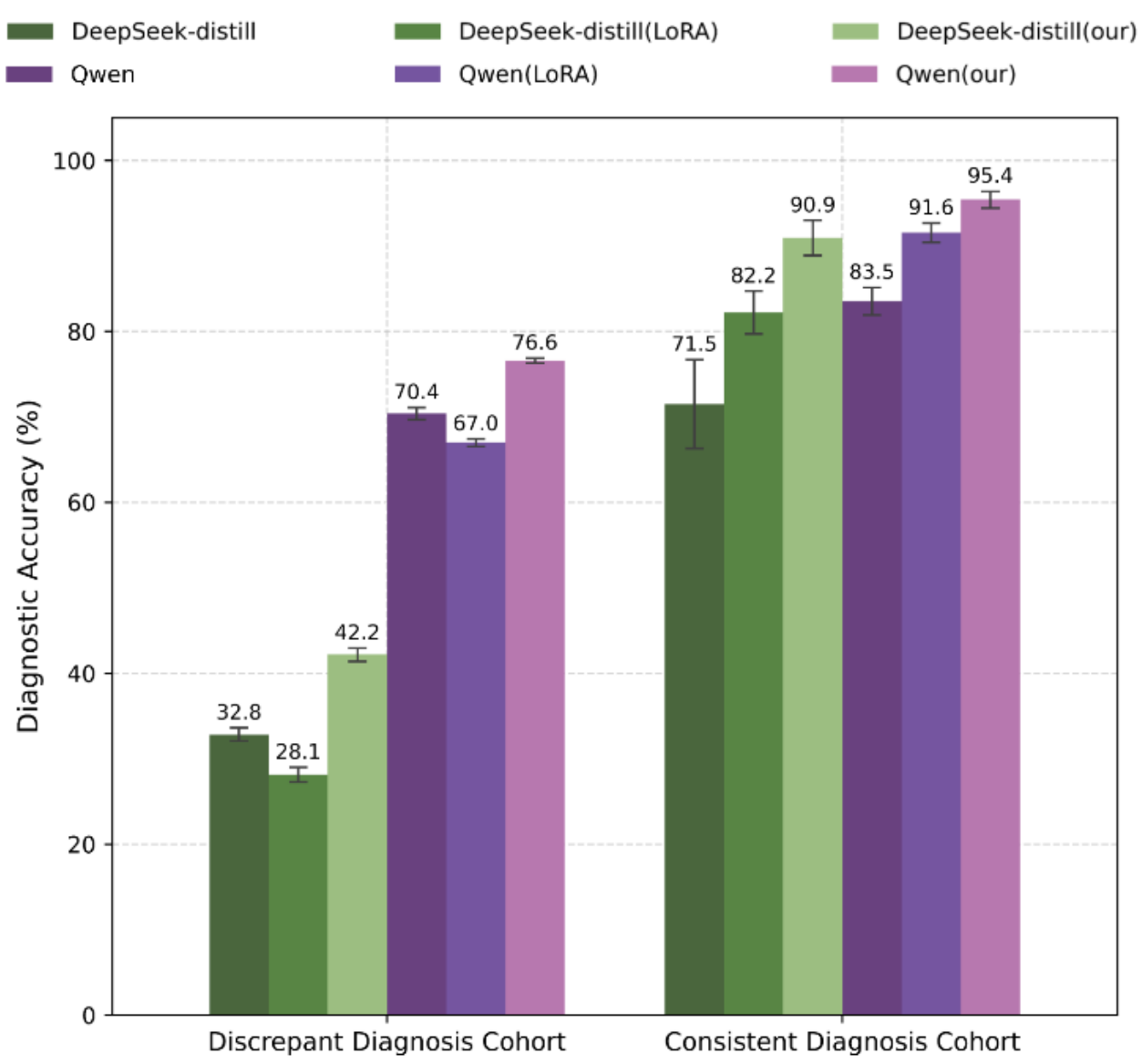

**Figure 7.** Diagnostic accuracy comparison across models in the Discrepant Diagnosis Cohort and Consistent Diagnosis Cohort. Both the LoRA and Ours models were fine-tuned on the Consistent Diagnosis Cohort using LoRA and structure-guided strategy,

respectively. For validation on the Discrepant Diagnosis Cohort, the best-performing folds from five-fold cross-validation were selected for each model.

To further investigate how different models attend to semantically meaningful information during diagnosis, we analyzed the Inference Attention Frequency on the Discrepant Diagnosis Cohort. Specifically, for each patient record, we extracted the top-10 most attended tokens during diagnostic generation and mapped them to their corresponding semantic segments. After filtering out trivial or non-informative segments, we computed the Inference Attention Frequency for each semantic segment, defined as the proportion of times a segment was attended to relative to its total occurrence in the evaluation dataset. The results are presented in Figure 8.

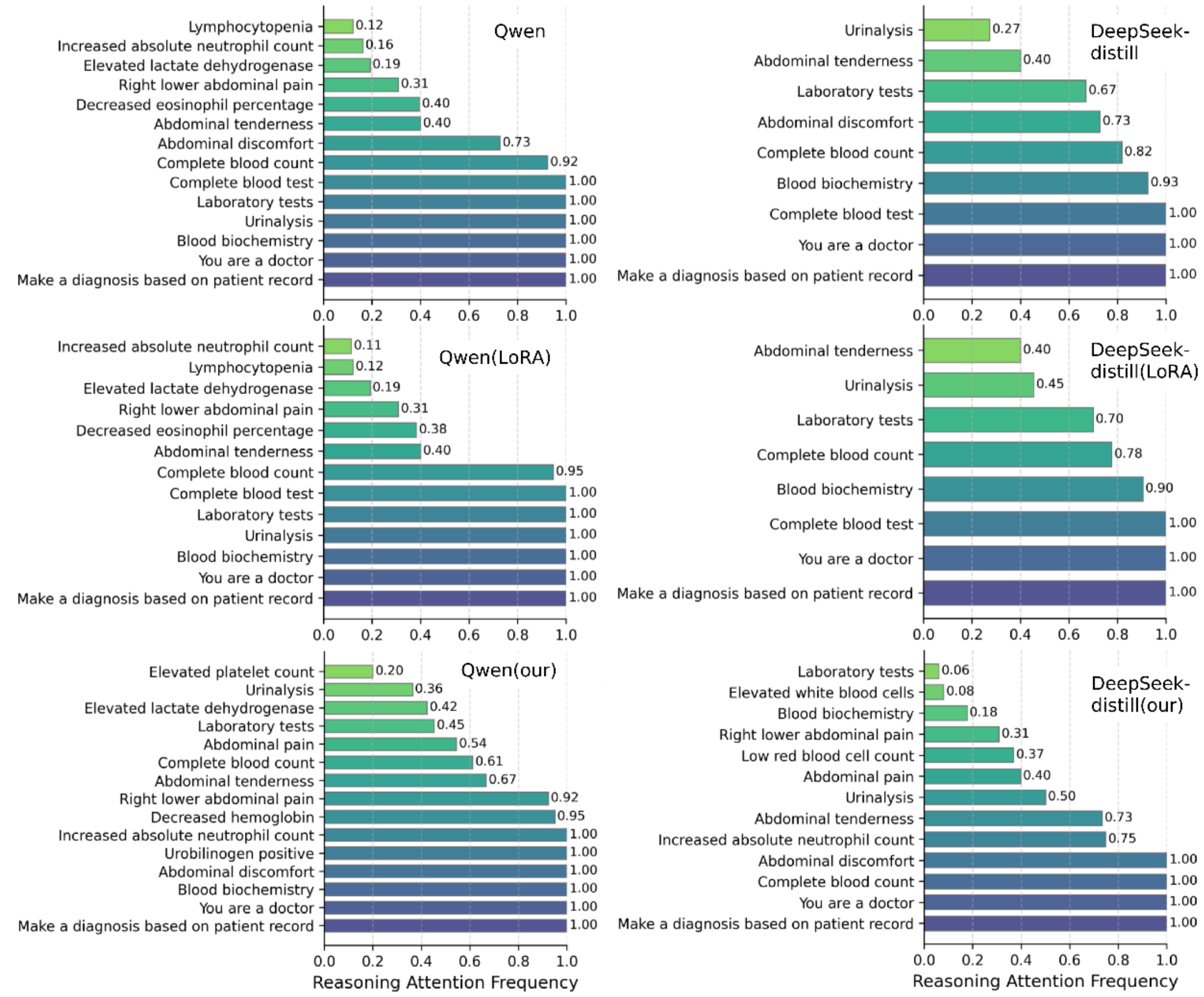

**Figure 8.** Inference Attention Frequency across models in the Discrepant Diagnosis Cohort. For each model, the top-10 attended semantic segments per sample were aggregated and normalized by their total occurrence in the dataset, yielding the Inference Attention Frequency that reflects how often models focus on clinically relevant information during diagnosis.

As illustrated, models fine-tuned with structure-guided attend to a richer and more diverse set of meaningful segments, achieving both a higher number of attended semantic segments and greater Inference Attention Frequency compared to both base models and LoRA-tuned models. Notably, the base models consistently outperform their LoRA counterparts in terms of attention to meaningful segments—both in quantity and frequency—mirroring the earlier observation that LoRA fine-tuning leads to performance degradation on the Discrepant Diagnosis Cohort. This finding suggests that conventional LoRA tuning may be insufficient to effectively redirect model attention toward clinically salient cues in reasoning-intensive diagnostic tasks.

In cross-model comparisons, Qwen-based models exhibit broader semantic coverage and higher Inference Attention Frequency than their DeepSeek-distill counterparts. This observation aligns with the more pronounced accuracy degradation observed for DeepSeek-distill on the Discrepant Diagnosis Cohort and indicates a lower robustness in capturing diagnostic cues under ambiguous or misleading input conditions.

## 5. Discussion

Evaluation on the Consistent Diagnosis Cohort revealed a clear performance stratification across disease complexity. While all baseline LLMs achieved high accuracy on acute appendicitis, performance degraded substantially for acute cholecystitis and acute pancreatitis—conditions that require integrating heterogeneous clinical signals rather than relying on single dominant features. This observation suggests that untuned general-purpose LLMs remain vulnerable to task settings where discriminative information is distributed across multiple clinical sources. Although parameter-efficient fine-tuning via LoRA improved overall accuracy, the gains were limited in these high-ambiguity conditions, indicating that LoRA primarily optimizes label fitting without reshaping how the model structurally processes etiological information. In contrast, the proposed structure-guided learning strategy consistently yielded superior and more balanced performance across all diseases. By explicitly aligning the model's attention with structured etiological cues during fine-tuning, the framework promotes more stable decision boundaries rather than superficial pattern adaptation. For example, Qwen(our) achieved a diagnostic accuracy of 96.6% for acute pancreatitis, substantially outperforming both its base and LoRA-finetuned counterparts. These results indicate that performance gains arise not from stronger generation capacity, but from more reliable and clinically grounded internal organization of diagnostically relevant information, laying the foundation for trustworthy diagnostic decision-making.

Insights into the internal mechanisms underlying these improvements were provided by the Etiology-Aware Head Identification analysis. In the Qwen model, several attention heads exhibited consistent sensitivity across multiple diagnostic stages, including physical examination, laboratory testing, and radiology reports. The high inter-stage and inter-disease overlap observed in the Jaccard similarity analysis suggests that these heads encode disease-agnostic etiological structures rather than task-specific surface patterns. In contrast, DeepSeek-distill demonstrated more heterogeneous attention distributions across diseases, implying greater reliance on context-dependent cues. These structural differences were reflected in downstream behavior: models benefiting from stable etiological attention patterns were more resilient to task complexity. Such stability at the attention level reflects a process-level form of trustworthiness, in which diagnostic decisions are supported by consistent utilization of clinically meaningful evidence rather than context-dependent surface cues. Importantly, improvements observed in R-Score and RougeL-F1 should not be interpreted as evidence of stronger abstract reasoning per se, but rather as indicators that structured supervision successfully reshaped attention allocation toward diagnostically meaningful input segments.

The ablation experiments further highlight the central role of structure-guided supervision. Removing Etiology-Aware Head Identification substantially degraded performance, indicating that indiscriminate attention intervention can disrupt established representational hierarchies. Similarly, removing the structure-guided loss while retaining CES annotations reduced performance, particularly for acute pancreatitis—the disease with the smallest sample size. This finding suggests that structure-guided attention supervision helps mitigate data imbalance by emphasizing fine-grained etiological distinctions rather than dominant class signals. Together, these results demonstrate that the effectiveness of the proposed framework depends on the coordinated interaction between structured annotations, selective attention targeting, and loss-level guidance, rather than any single component in isolation.

The Discrepant Diagnosis Cohort provides a more stringent test of diagnostic trustworthiness under realistic clinical conditions characterized by incomplete and misleading information. In this setting, LoRA fine-tuning consistently led to performance degradation, suggesting sensitivity to distributional shifts between training and evaluation data. In contrast, the proposed structure-guided fine-tuning approach maintained stable accuracy for Qwen and produced substantial gains for DeepSeek-distill. These results indicate that explicitly guiding attention toward etiologically grounded information enables the model to preserve discriminative capacity even when superficial correlations fail. Rather than serving as external validation in the strict sense, this cohort demonstrates the framework's ability to stabilize diagnostic behavior under clinically adverse conditions.

Further evidence for this stabilization effect is provided by the Inference Attention Frequency analysis. While all models appropriately attended to task instructions, only the structure-guided fine-tuning models consistently prioritized clinically informative segments such as localized pain descriptions and abnormal laboratory findings. In contrast, base and LoRA-tuned models disproportionately focused on generic but weakly discriminative terms. Notably, despite being fine-tuned using English-language CES annotations, the models exhibited coherent attention alignment on Chinese clinical records, suggesting that the supervision operated at a semantic and structural level rather than through language-specific token memorization. This cross-lingual alignment further supports the claim that structure-guided attention supervision enhances general sensitivity to clinically meaningful content rather than inducing brittle pattern dependence.

Despite these encouraging findings, several limitations remain. The CES was manually constructed based on clinical guidelines and relied on selected attention heads, which may constrain scalability to broader disease categories. Additionally, the identification of etiological elements followed heuristic rules and lacked dynamic adaptation, potentially limiting flexibility in atypical or rare presentations. Addressing these limitations will require more automated and adaptive mechanisms for structure discovery.

Nevertheless, this work demonstrates that explicitly incorporating structured etiological knowledge into attention supervision offers a viable pathway for improving the trustworthiness of large language models in complex clinical diagnostic scenarios, by aligning internal attention mechanisms with structured, clinically meaningful etiological evidence. By shifting fine-tuning from label-centric optimization toward structure-guided learning, the proposed Etiology-Aware Attention Supervision framework bridges the gap between diagnosis generation and clinically grounded information processing. The resulting models exhibit improved accuracy, greater stability under distributional shifts, and more interpretable attention behavior, highlighting the potential of structure-aware fine-tuning strategies for real-world clinical deployment.

By shifting supervision from output-level correctness to process-level evidence utilization, this work contributes toward the development of clinically trustworthy diagnostic language models, whose decisions can be both effective and auditable. Future work may explore automated discovery of structures using explainability or reinforcement learning techniques, extend the framework to multi-disease and multimodal settings, and integrate self-supervised or contrastive objectives to further enhance robustness under noisy or low-resource conditions.

## 6. Conclusion

This study proposes an Etiology-Aware Attention Supervision framework to enhance the trustworthiness of large language models in clinical diagnostic decision-making under complex and uncertain scenarios, by introducing structured, clinically grounded etiological information into parameter-efficient fine-tuning. By identifying attention heads consistently aligned with etiological elements and steering attention behaviors through structure-guided supervision during training, the proposed method reshapes internal attention organization beyond label-level optimization, enabling clinically consistent, interpretable, and inspectable diagnostic decision processes. Overall, this study establishes a robust and trustworthy pathway for deploying medical LLMs in high-risk diagnostic settings.

## Declaration of competing interest

The authors declare that they have no known competing financial interests or personal relationships that could have appeared to influence the work reported in this paper.

## CRediT author statement

Peixian Li: Investigation, Conceptualization, Methodology, Software, Writing – original draft, Writing – review & editing. Yu Tian: Investigation, Conceptualization, Methodology, Writing – review & editing, Funding acquisition, Project administration. Ruiqi Tu: Software. Chengkai Wu: Methodology. Jingjing Ren: Resources, Data Curation. Jingsong Li: Supervision, Writing – review & editing, Funding acquisition.

## Data availability

The Consistent Diagnosis Cohort is derived from the publicly available MIMIC-IV database. The Discrepant Diagnosis Cohort contains non-public clinical data with data-sharing restrictions.

## Acknowledgments

This work was supported by the National Natural Science Foundation of China (No. 82572379, No. 62402455, No. 72274169), the Major Research Plan of the National Natural Science Foundation of China (No. 92474301), Fundamental Research Funds for the Central Universities (No. 226-2025-00006), Innovation and Development Special Fund of Hangzhou West Sci-Tech Innovation Corridor.